\documentclass[letterpaper]{article} 
\usepackage{aaai25}  
\usepackage{times}  
\usepackage{helvet}  
\usepackage{courier}  
\usepackage[hyphens]{url}  
\usepackage{graphicx} 
\usepackage{natbib}  
\usepackage{caption} 
\usepackage{algorithm}
\usepackage{algorithmic}

\usepackage{newfloat}
\usepackage{listings}
\DeclareCaptionStyle{ruled}{labelfont=normalfont,labelsep=colon,strut=off} 
\floatstyle{ruled}
\newfloat{listing}{tb}{lst}{}
\floatname{listing}{Listing}
\usepackage{amsmath}
\usepackage{amsthm}
\usepackage{amssymb}
\usepackage{booktabs}
\usepackage{subcaption}

\title{Learning Visually Grounded Domain Ontologies\\ via Embodied Conversation and Explanation}
\author{
    Jonghyuk Park,
    Alex Lascarides,
    Subramanian Ramamoorthy
}
\affiliations{
    School of Informatics, The University of Edinburgh\\

    \{jay.jh.park, a.lascarides, s.ramamoorthy\}@ed.ac.uk
}

\usepackage{bibentry}

\begin{document}

\maketitle

\begin{abstract}
In this paper, we offer a learning framework in which the agent's knowledge gaps are overcome through corrective feedback from a teacher whenever the agent explains its (incorrect) predictions.
We test it in a low-resource visual processing scenario, in which the agent must learn to recognize distinct types of toy truck.
The agent starts the learning process with no ontology about what types of trucks exist nor which parts they have, and a deficient model for recognizing those parts from visual input.
The teacher's feedback to the agent's explanations addresses its lack of relevant knowledge in the ontology via a generic rule (e.g., ``dump trucks have dumpers''), whereas an inaccurate part recognition is corrected by a deictic statement (e.g., ``this is not a dumper'').
The learner utilizes this feedback not only to improve its estimate of the hypothesis space of possible domain ontologies and probability distributions over them, but also to use those estimates to update its visual interpretation of the scene.
Our experiments demonstrate that teacher-learner pairs utilizing explanations and corrections are more data-efficient than those without such a faculty.
\end{abstract}

%
\begin{links}
    \link{Code}{https://github.com/jpstyle/ns-arch-unity}
\end{links}

\section{Introduction}

The field of eXplainable Artificial Intelligence (XAI) aims to develop AI systems that can explain their decisions, such that any identified knowledge gaps can be closed to improve performance on various dimensions: e.g., task success rates, faster convergence, or closer alignment with human knowledge.
However, many XAI works focus only on identifying knowledge gaps, not exploiting those gaps in a principled way as learning opportunities \cite{weber2023beyond}.
Furthermore, the currently most popular ML models that require explanations are deep neural networks (DNNs), which generally need many iterations of parameter updates to incur significant behavior changes.
Providing human feedback to every single model explanation would be infeasible for DNNs, so most existing work on XAI-based model improvements aims to involve as little human involvement as possible during training.

\begin{figure}
\centering
\begin{subfigure}{0.45\textwidth}
    \includegraphics[width=\textwidth]{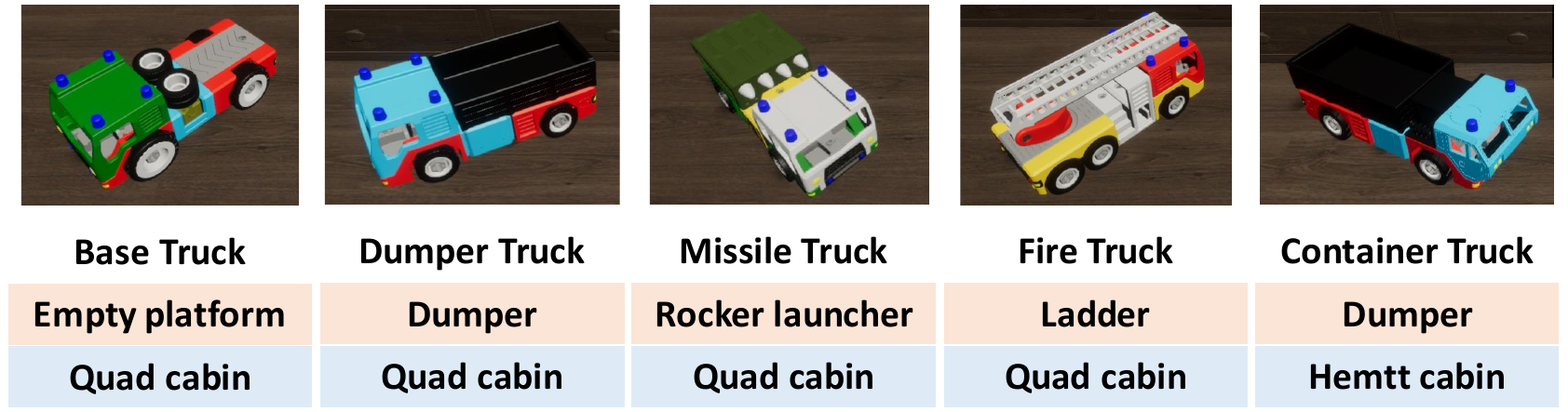}
    \caption{3D models of toy trucks characterized by parts.}
    \label{fig:truck_domain}
\end{subfigure}
\par
\begin{subfigure}{0.45\textwidth}
    \includegraphics[width=\textwidth]{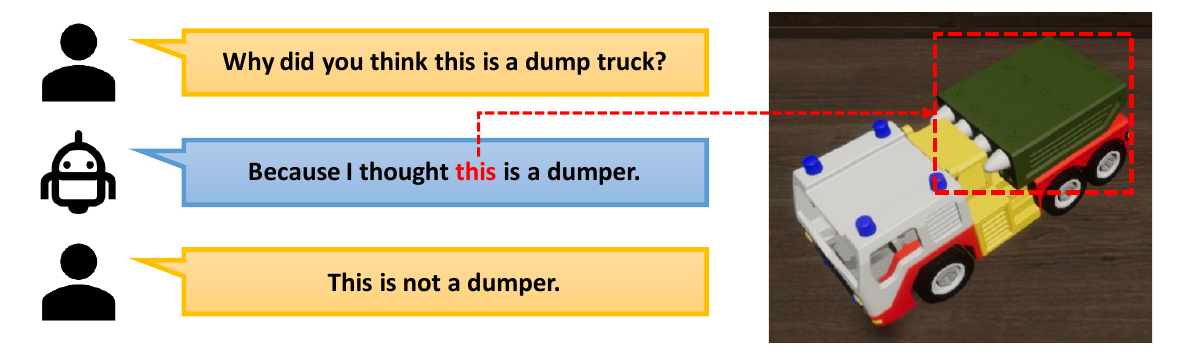}
    \caption{Example interaction between a teacher and a learner featuring a part-based explanation and corrective feedback.}
    \label{fig:opening_ex}
\end{subfigure}
\caption{Explanatory interactive learning in a simulated domain.}
\end{figure}

In the meantime, feedback by a human-in-the-loop in response to XAI explanations could prove much more valuable when the learner has to quickly adapt to an unfamiliar domain with a continuously changing hypothesis space.
Consider, for example, a general-purpose robotic agent deployed in an assembly plant where task concepts may be frequently added or modified in unforeseen ways, and supervision by a human domain expert (perhaps without ML expertise) is available via natural dialogues.
Concepts in the domain ontology may be niche and specific to this particular plant, making it difficult to address the learner's initial ignorance of the domain ontology with some existing pre-trained ML models.
In such scenarios, it would be desirable for the expert's corrective feedback to the agent's explanations (as demonstrated in Fig.~\ref{fig:opening_ex}) to cause immediate behaviour changes without interrupting its operation, so as not to exhaust or irritate the human teacher.
This motivation is particularly pertinent when dealing with highly specialized, low-resource domains, where it is difficult to prepare automated feedback to agent explanations.

In this paper, we propose a framework for explanatory interactive learning (XIL; cf.~\citealt{teso2023leveraging}), suited to a family of knowledge-based neurosymbolic AI systems.
The primary strength of our approach is that the learner can start out with scarce symbolic knowledge of the domain ontology, completely unaware of many crucial concepts, let alone their relationships to each other.  It learns online and incrementally through continued natural interaction with a human supervisor.
This contrasts with a nontrivial assumption commonly made in most neurosymbolic architectures proposed so far, that the symbolic domain knowledge is either fully encoded or batch-learned before deployment.

We instantiate the framework in an example testbed scenario where a learner agent must acquire and distinguish among a set of novel visual concepts, namely a range of fine-grained types of toy vehicle.
We empirically demonstrate how the learner's explanations can induce tailored feedback from the teacher that addresses the exposed knowledge gaps, leading to two distinct types of knowledge update: 1) acquisition of generic rule-like knowledge about `have-a' relations between whole and part concepts; and 2) joint refinement of recognition capabilities of visual concepts, for both object wholes and their parts.
Our experiments demonstrate that strategies exploiting agent explanations in this way accomplish significantly better performance after the same number of training examples, compared to baseline strategies that do not---especially when the learner's initial model for recognizing object parts is deficient.

\section{Explanatory Interactive Learning}
\label{sec:framework}

\subsection{Testbed task: fine-grained visual classification}

The main challenge of fine-grained visual classification (FGVC) is that instances of fine-grained subcategories exhibit small inter-class variance and large intra-class variance \cite{wei2021fine}.
Recent development of powerful vision foundation models has enabled strong FGVC performance with only visual features and label annotations, adding a classification head on top of the pre-trained models with weights frozen \cite{oquab2023dinov2}, which we will employ as a `vision-only' baseline in this study.

Formally, each input to our FGVC task is a pair $x_i=(\mathcal{I}_i,m_i)$ with an RGB image $\mathcal{I}_i$ and a binary mask $m_i$.
Thus, $x_i$ essentially refers to an object in an image.
The anticipated task output $y_i\in C_\text{fg\_type}$ is the correct concept of the object instance $x_i$, where $C_\text{fg\_type}$ is a set of fine-grained visual concepts.
One important difference between our problem setting versus traditional FGVC is that for us $C_\text{fg\_type}$ is not known to the learner at the start---it must be acquired through interaction with the teacher.
The classification target concepts in our experiments are fine-grained types of toy truck as illustrated in Fig.~\ref{fig:truck_domain}.
Some features define the fine-grained types of truck and thus are crucial for their successful classification (types of load and cabin parts), while other irrelevant features may vary across the types and act as distracting signals (e.g., colours of different body parts, numbers and sizes of wheels).
The learner starts out knowing that toy trucks have loads and cabins, albeit with imperfect recognition capability.

\subsection{Dialogue Strategies}
\label{sec:framework:dialogue}

\begin{figure}[t]
\centering
\includegraphics[width=0.45\textwidth]{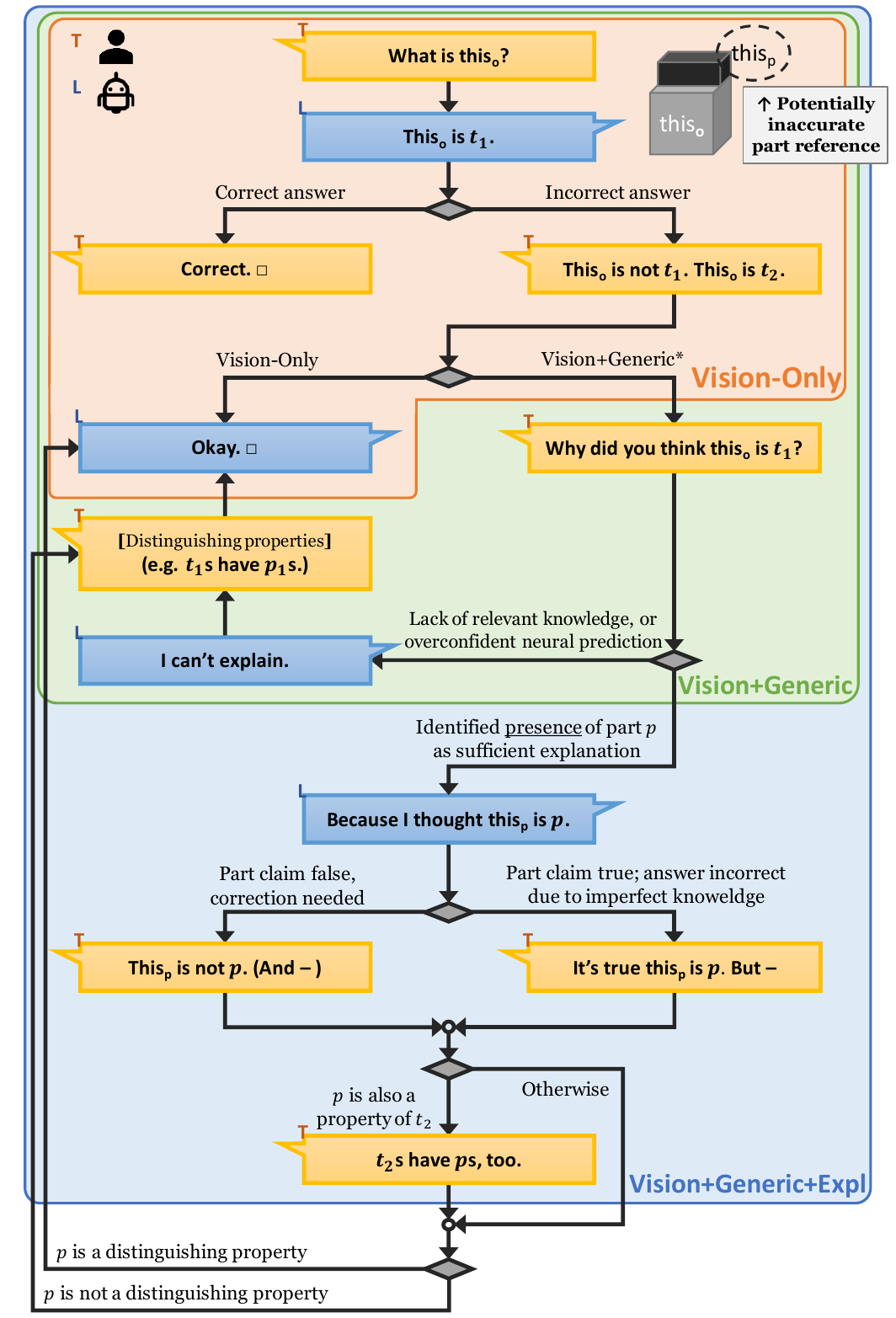}
\caption{Flowchart covering the range of training dialogues modeled in this study. $\square$ marks termination of an episode.}
\label{fig:dialogue}
\end{figure}

Fig.~\ref{fig:dialogue} depicts dialogue flows that we aim to cover with our proposed approach and baselines.
In effect, each episode of interaction is a single round of FGVC inference and learning with a training example, unrolled into a sequence of NL utterances exchanged between the teacher and the learner.
An interaction episode starts with the teacher presenting an instance $o$ of one of the target visual concepts in $C_\text{fg\_type}$.
The teacher asks a probing question ``What kind of truck is this$_o$?'' along with an accurate binary mask specifying the object demonstratively referenced by ``this$_o$''.
The learner answers with its best estimate of the fine-grained type in $C_\text{fg\_type}$; e.g., ``This$_o$ is a dump truck''.
This answer may be right or wrong; if the object is indeed a dump truck, then the episode terminates, and the teacher and learner proceed to the next episode with a new example.
If the answer is wrong, then the teacher corrects the learner's answer by uttering, for instance, ``This$_o$ is not a dump truck. This$_o$ is a missile truck''.

After the teacher's correction, the dialogue branches according to whether the learner relies on direct neural prediction only or utilizes generic knowledge about distinctive part properties as well.
The former case is a baseline, which we refer to as \textbf{Vis-Only}.
No further interactions ensue, and the episode terminates.
Otherwise, the teacher investigates the source of the error by asking ``Why did you think this$_o$ is a dump truck?''.
This introduces another branching point that hinges upon whether the learner is able to explain the most relevant reason why it concluded so.
We refer to the configuration where the learner lacks such a faculty as the \textbf{Vis+Genr} baseline, and the opposite case with the complete toolkit (i.e., our full approach) as \textbf{Vis+Genr+Expl}.
There are two alternative scenarios in which the learner answers ``I cannot explain'' to the teacher's \textit{why}-question.
Either:
\begin{enumerate}
    \item The episode takes place under \textbf{Vis+Genr} setting; or
    \item The episode takes place under \textbf{Vis+Genr+Expl} setting, but the learner lacks relevant generic rules about whole-part relations in its KB, or the vision module made an overconfident (but wrong) prediction on the whole in spite of not recognizing its parts.
\end{enumerate}
The teacher responds to ``I cannot explain'' with rule-like knowledge about properties that separate the truck types: e.g., ``Missile trucks have rocket launchers, and dump trucks have dumpers''.
The episode then terminates.

If the learner has leveraged recognition of some truck part as evidence, the learner reports a part-based explanation such as ``Because I thought this$_p$ is a dumper'' while accompanying the demonstrative ``this$_p$'' with a pointing action that corresponds to its estimated binary mask.
Note that this explanation implies the learner is aware of the rule ``dump trucks have dumpers'' and recognizes having dumpers as a distinguishing property.

There are various potential sources of error that can be exposed by this explanation.
If the object referenced by ``this$_p$'' is not a dumper as claimed, or the estimated mask has poor quality and fails to specify an object at all, the explanation itself needs correction, and the teacher says ``This$_p$ is not a dumper''.
The learner can then memorize the quoted part as a negative exemplar of the dumper concept.
Alternatively, the part claim may be correct in its own right, but dumpers may not be a distinguishing feature between the learner's estimate concept and the ground truth concept.
For instance, in the domain illustrated by Fig.~\ref{fig:truck_domain}, dumper trucks and container trucks both have dumpers but have different type of cabins: quad vs. hemtt.
In this case, the teacher first acknowledges ``It's true that this$_p$ is a dumper'', then corrects the learner's reasoning by stating ``But container trucks have dumpers, too'', effectively depriving the `dumper' concept of the role as a distinguishing property.
Since the part cited as explanation is not a distinguishing property between the agent's answer vs. the true answer, it is implied that the learner may be ignorant of the true distinguishing feature.
So the teacher provides the gap in the learner's knowledge via a further generic utterance: e.g., ``Dumper trucks have quad cabins while container trucks have hemtt cabins''.
The episode then terminates.

\section{Neurosymbolic Agent Architecture}
\label{sec:architecture}

\begin{figure}
\centering
\includegraphics[width=0.35\textwidth]{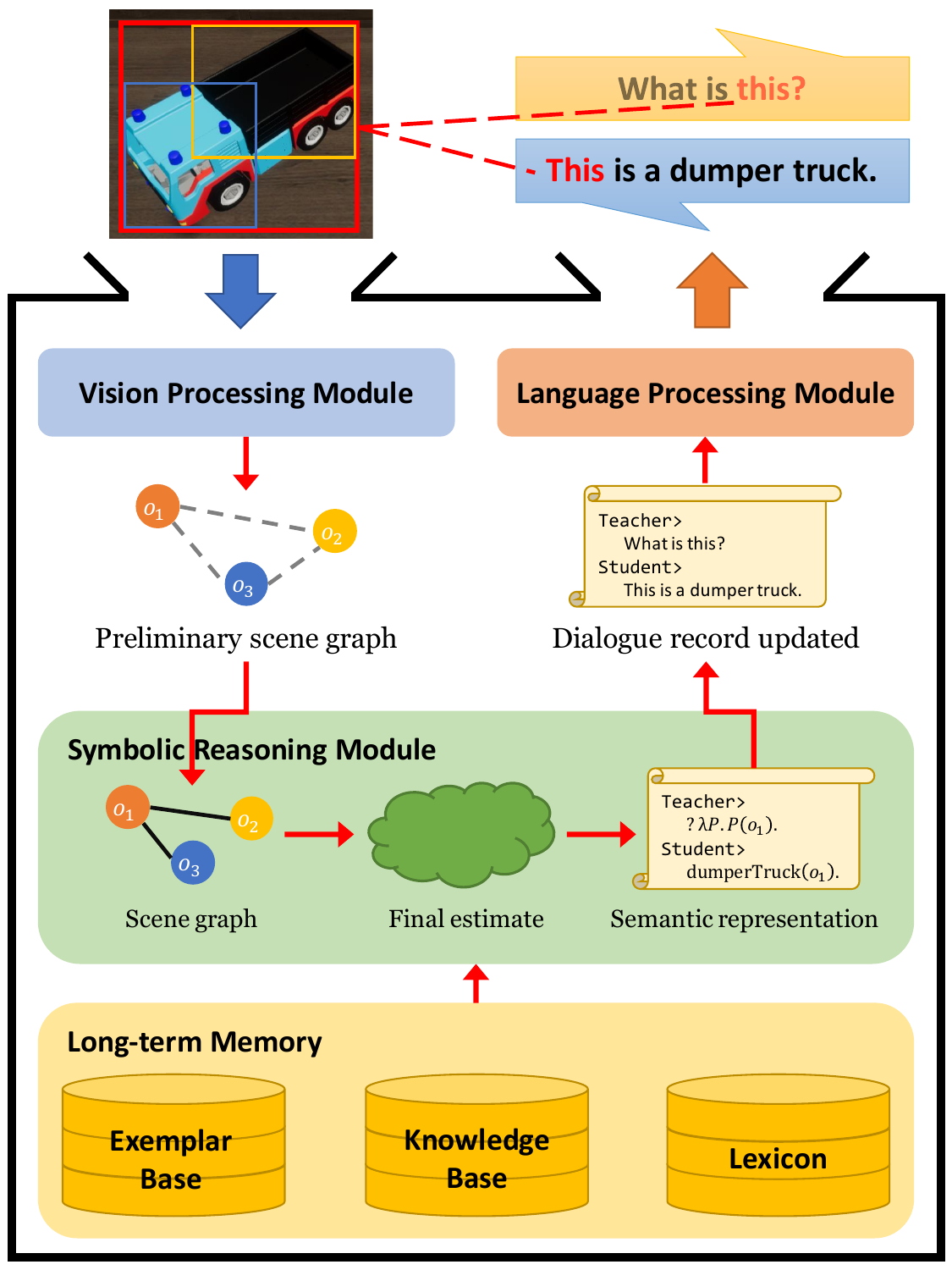}
\caption{Overview of the proposed neurosymbolic architecture in inference mode, in which the component modules interact to generate an answer to a user question.}
\label{fig:arch}
\end{figure}

This section outlines our neurosymbolic architecture, assuming it is deployed to learn and perform image analysis tasks.
However, note that in principle our framework is designed for knowledge-intensive tasks with any input modality (image, video, language, audio, etc.) where dialogue participants can naturally refer to parts of inputs.
Fig.~\ref{fig:arch} illustrates the system components and their coordination in inference mode.
As depicted, the agent architecture has four components with distinct responsibilities: vision processing, language processing, long-term memory and symbolic reasoning.
The architecture is highly modular and can employ any existing methods for each component as long as they satisfy the feature requirements described in the subsections below.

\subsection{The vision processing module}
\label{sec:arch:vision}

The vision processing module is responsible for parsing raw pixel-level inputs perceived by the system's vision sensor into structured representations.
We use an object-centric graph-like data structure for summarizing visual scenes, which we will refer to as \textit{scene graphs} hereafter.
Scene graphs encode a set of salient objects in a visual scene and their conceptual relations (e.g., $dumpTruck(o_i)$, $have(o_i,o_j)$) along with likelihood scores.
Each object in a scene graph is also associated with a demonstrative reference to it by segmentation mask.
A scene graph serves as an internal, abstract representation of an image that will be later processed by symbolic reasoning methods.

The functional requirement we impose on the vision processing module is twofold.
First, given a raw RGB image and a set of $n$ binary masks $\{m_1,\cdots,m_n\}$ referring to scene objects, the vision module should be able to perform few-shot classification on visual concepts it's aware of, returning the estimated probabilities of concept membership as its output.
Second, given a raw RGB image and a visual concept $\gamma$, the vision module should be able to localize and segment instances of $\gamma$ in the visual scene in few shot (i.e., $\gamma$ is specified by a moderately sized set of positive exemplars $\chi_\gamma^+$), producing a set of corresponding binary masks $\{m_1,\cdots,m_p\}$ for some $p$.
This `visual search' functionality is needed for recognizing object parts as evidence that would significantly affect later symbolic reasoning.
For instance, if it is known that existence of a dumper on a truck will significantly raise the probability that the truck is a dumper truck, instances of dumper should be searched for if not already included in the scene graph due to being less salient.

We will refer to the classification and segmentation functions defined for a concept $\gamma$ as $f_\text{clf}^\gamma$ and $f_\text{seg}^\gamma$ hereafter.
It is important that both $f_\text{clf}^\gamma$ and $f_\text{seg}^\gamma$ can operate few-shot, so as to enable quick, online acquisition of novel domain concepts during operation; the learner shouldn't require several thousands of annotated examples from the human user to acquire unforeseen domain concepts.
We also need to ensure that $f_\text{clf}^\gamma$ and $f_\text{seg}^\gamma$ perform better as exemplar sets grow larger with continued interaction with the human user, in order to allow gradual improvement of object recognition from visual data.

Refer to the Technical Appendix for a more formal, detailed description of how scene graphs are represented.

\subsection{The language processing module}
\label{sec:arch:lang}

The language processing module handles NL interactions with human users.
Its core responsibilities include:
\begin{itemize}
    \item Parsing free-form NL inputs from users into their semantic contents and pragmatic intents (natural language understanding; NLU).
    \item Incrementally updating semantic representations of states of ongoing dialogues and deciding which speech acts to perform from current states (dialogue management; DM).
    \item Generating NL utterances according to the DM's decisions about the agent's next speech act (natural language generation; NLG).
\end{itemize}
We draw on formal semantic representations for the NLU, DM and NLG components.
In the scope of this work, each clause in the dialogues is either a proposition or a question.
Suppose we have a standard first-order language $\mathcal{L}$ defined for describing scene graphs, including constants referring to objects and predicates referring to visual concepts.
We will represent the semantic content of an indicative NL sentence with an antecedent-consequent pair of $\mathcal{L}$-formulae; we will refer to such a unit as a \textsc{prop} hereafter.
If $\psi$ is a \textsc{prop} comprising an antecedent $\mathcal{L}$-formula $Ante$ and a consequent $\mathcal{L}$-formula $Cons$, then $\psi$ has the form $Ante\Rightarrow Cons$.
We refer to $Ante$ and $Cons$ of $\psi$ as $Ante(\psi)$ and $Cons(\psi)$ respectively.
When a \textsc{prop} $\psi$ is a non-conditional, factual statement, $Ante(\psi)$ is empty and we omit the $Ante\Rightarrow$ part, leaving $Cons$.
For example, the NL sentence ``$o$ is a dump truck'' is represented as:
\begin{equation} \label{eq:fact_prop}
    dumpTruck(o)
\end{equation}
Quantified \textsc{prop}s are marked with matching quantifiers.
In this work, we are primarily concerned with generic statements and will let $\mathbb{G}$ denote the generic quantifier.
Thus, for example, the NL sentence ``Dump trucks (generally) have dumpers'' can be represented as follows:
\begin{equation}
    \mathbb{G}x.dumpTruck(x)\Rightarrow (\exists y.have(x,y)\land dumper(y))
\end{equation}
However, we go further and use a skolem function $f$ in $\mathcal{L}$ that maps an instance of dump truck to its dumper:
\begin{equation} \label{eq:genr_prop}
    \mathbb{G}x.dumpTruck(x)\Rightarrow have(x,f(x))\land dumper(f(x))
\end{equation}
(\ref{eq:genr_prop}) is more compatible with the logic programming formalism that we adopt in the symbolic reasoning module.

Our semantic representation of questions, \textsc{ques} hereafter, resembles the notation from \citet{groenendijk1982semantic}.
A \textit{wh}-question is represented as $?\lambda x.\psi(x)$ where $\psi$ is a \textsc{prop} in which the variable $x$ is free.
It's (partial) answers provide values $a$ for $x$ such that $\psi[x/a]$ evaluates to true.
For example, we represent the NL sentence ``What kind of truck is $o$?'' by the following \textsc{ques}:
\begin{equation} \label{eq:type_ques}
    ?\lambda P.P(o)\land\vdash^{type}_{\mathit{KB}}(P,truck)
\end{equation}
by taking the liberty of interpreting the NL question as meaning ``Which concept $P$ has $o$ as an instance and entails that $o$ is a truck?''.
Here, $\vdash^{type}_{\mathit{KB}}$ is a reserved predicate such that $\vdash^{type}_{\mathit{KB}}(p_1,p_2)$ evaluates to true if and only if $p_1$ is a subtype of $p_2$ according to the domain ontology represented by a knowledge base $KB$.

Another important type of question in the interactions from Fig.~\ref{fig:dialogue} is \textit{why}-questions.
We simply introduce another symbol $?why_{\mathit{DP}}$ for representing \textit{why}-questions that query the source of belief by some dialogue participant $\mathit{DP}$.
For example, we represent the NL question ``Why did you think $o$ is a dump truck?'' addressed to the learner agent $agt$ as:
\begin{equation} \label{eq:why_ques}
    ?why_{agt}.dumpTruck(o)
\end{equation}

\subsection{The long-term memory module}
\label{sec:arch:lt_mem}

The long-term memory module stores interpretable fragments of knowledge, which the learner has acquired from the teacher over the course of interaction.
Three types of storage subcomponents are needed: the visual exemplar base (XB), the symbolic knowledge base (KB) and the lexicon.

The visual XB stores collections of positive and negative concept exemplars $\chi_\gamma^+$ and $\chi_\gamma^-$ for each visual concept $\gamma$.
$\chi_\gamma^+$ and $\chi_\gamma^-$ naturally induce a binary classifier for the concept $\gamma$, which is used by $f_\text{clf}^\gamma$ in the vision module.
$\chi_\gamma^+$ and $\chi_\gamma^-$ are updated whenever the learner encounters an object and deems it worth remembering as a positive/negative sample of $\gamma$.
Whenever the learner encounters an unforeseen concept $\gamma$ via the teacher's utterance that features a word that's outside the learner's vocabulary (i.e., a neologism), it is registered as a new concept, and new empty sets $\chi_\gamma^+$ and $\chi_\gamma^-$ are created in the XB.

The symbolic KB stores a collection of generic rules about relations between concepts, represented as $\mathbb{G}$-quantified \textsc{prop}s like (\ref{eq:genr_prop}).
Generic rules are acquired via interactions with a teacher and stored in the KB.
In our setting, the generic statements express `have-a' relations between whole vs. part concepts.

The lexicon stores a mapping between visual concepts and NL content word strings.
Neologisms referring to novel visual concepts are added to the lexicon when first mentioned by a teacher, accordingly updating the visual concept vocabulary and the XB.

\subsection{Symbolic reasoning module}
\label{sec:arch:reasoning}

The symbolic reasoning module combines subsymbolic perceptual inputs and symbolic relational knowledge.
The `first impressions' formed by the vision module are adjusted by the generic rules in KB to yield final estimations of the current image's interpretation.
To cope with estimation of quantitative uncertainties on logical grounds, we use probabilistic graphical models for symbolic reasoning.

\begin{figure}[t]
\centering
\includegraphics[width=0.45\textwidth]{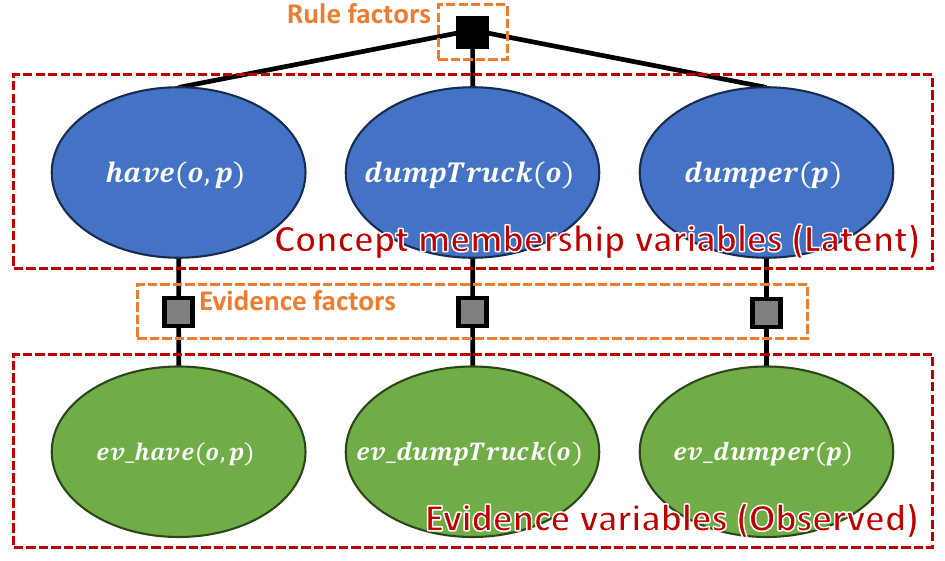}
\caption{An example truck classification problem represented as a probabilistic factor graph model.}
\label{fig:gmodel}
\end{figure}

Fig.~\ref{fig:gmodel} illustrates a small example of how we represent an instance of a visual classification problem as a factor graph model.
Our factor graph representations feature two types of variable nodes and two types of factor nodes.
Concept membership variables represent whether an object (or an ordered tuple of objects) is `truly' an instance of a concept.
Evidence variables represent evidence with quantitative uncertainty, obtained through a process external to the scope of the reasoning problem (\textit{virtual evidence}; cf. \citealt{pearl1988probabilistic}), namely visual perception and neural prediction here.
Each evidence variable is uniquely defined for and causally linked to the corresponding concept membership variable.
Evidence factors connect each matching pair of a concept membership variable and an evidence variable.
The strengths of observed evidence, which are exactly the probability values returned by $f_\text{clf}^\gamma$ functions, are entered as input to corresponding evidence factors as likelihood ratios: e.g., $\text{Pr}(ev\_dumper(p)=T|dumper(p)=T):\text{Pr}(ev\_dumper(p)=T|dumper(p)=F)$.
Rule factors connect concept membership variables and represent KB entries that encode relational knowledge like (\ref{eq:genr_prop}).

We construct a graphical model for inference by first translating the scene graph and the symbolic KB entries into a probability-weighted normal logic program.
The primary reason we employ normal logic programs as an intermediate representation is that we would like the generic \textsc{prop}s in the KB to receive minimal model interpretations.
In other words, we want to infer a rule consequent $Cons$ only when some rule antecedent $Ante$ that entails $Cons$ is proven to hold, as opposed to material implication in propositional logic and Markov logic \cite{richardson2006markov}.
This enables inference in the \textit{abductive} direction, e.g., raising the probability of an object being a dump truck after recognizing a dumper as its part due to the rule ``Dump trucks have dumpers''.
Default negations used in normal logic programs would also facilitate better treatment of the generic quantifier $\mathbb{G}$ through non-monotonic reasoning, though such use is outside the scope of this work.

Likelihood values encoded in a scene graph are compiled into a program $\Pi_V$ that summarizes visual evidence.
For instance, the likelihood that an object $o$ is an instance of a concept $\gamma$ is converted into a fragment of $\Pi_V$ as follows:
\vspace{1.5mm}
\begin{lstlisting}[frame=single,mathescape=true,numbers=none]
$0.5$ $::$ $\gamma(o)$.
$p$ $::$ $ev\_\gamma(o)\leftarrow \gamma(o)$.
$1-p$ $::$ $ev\_\gamma(o)\leftarrow \text{not }\gamma(o)$.
\end{lstlisting}
\vspace{1.5mm}
The number annotated to the left of each rule indicates that the rule head can hold with probability equal to that value if and only if the rule body is satisfied.
Note how the translation captures a Bayesian viewpoint; we can interpret that 0.5 in the first line represents a uniform prior, and $p$ and $1-p$ (where $p$ is obtained from $f_\text{clf}^\gamma$) represent data likelihoods.

The symbolic KB is translated into a program $\Pi_K$ that implements logical inference in both deductive and abductive directions.
For deductive inference, each KB entry $\kappa$ adds to $\Pi_K$ the following rule parametrized by $U_d\in [0,1]$:
\vspace{1.5mm}
\begin{lstlisting}[frame=single,mathescape=true,numbers=none]
$U_d$ $::$ $\leftarrow Ante(\kappa),\text{ not }Cons(\kappa)$.
\end{lstlisting}
\vspace{1.5mm}
which penalizes `deductive violation' of the entry $\kappa$.
Here, a rule without a head represents an integrity constraint, as is standard in normal logic programs.
For abductive inference, each set of KB entries $\{\kappa_i\}$ that share identical $Cons$ adds to $\Pi_K$ the following rule parametrized by $U_a\in [0,1]$:
\vspace{1.5mm}
\begin{lstlisting}[frame=single,mathescape=true,numbers=none]
$U_a$ $::$ $\leftarrow Cons, \bigwedge_{\{\kappa_i|Cons(\kappa_i)=Cons\}}\{\text{not }Ante(\kappa_i)\}$.
\end{lstlisting}
\vspace{1.5mm}
which penalizes failure to explain $Cons$ observed.
The parameters $U_d$ and $U_a$ can be understood as encoding the extent to which the agent relies on its symbolic knowledge.
We fix $U_d=U_a=0.99$ in our experiments.

After $\Pi_V$ and $\Pi_K$ are obtained, the program $\Pi_V\cup\Pi_K$ is translated into an equivalent factor graph.
For instance, having (\ref{eq:genr_prop}) as a KB entry will add the subgraph shown in Fig.~\ref{fig:gmodel} to the factor graph.
The symbolic reasoning module needs to be equipped with an inference algorithm for probabilistic graphical models, which should be implemented so as to correctly accommodate the semantics of normal logic programs.

\subsection{The agent's inference and learning procedures}
\label{sec:arch:procedures}

We now describe how the architecture components coordinate in service to the four cognitive processes integral to our framework: inference, explanation, instance-level learning and rule-level learning.

\subsubsection{Inference}

The agent performs FGVC inference upon parsing a probing \textsc{ques} like (\ref{eq:type_ques}) from the teacher.
The vision module first prepares a scene graph that includes the classification target object $o$ (referenced as ``this$_o$'' in Fig.~\ref{fig:dialogue}).
If the symbolic KB is not empty, potential instances of relevant part subtypes as determined by the KB are searched by $f_\text{seg}^\gamma$ and added to the scene graph as well.
Then, each concept membership probability for the scene objects are estimated by $f_\text{clf}^\gamma$, completing the scene graph $SG$.
$SG$ and the symbolic KB are each translated into probability-weighted normal logic programs $\Pi_V$ and $\Pi_K$ respectively.
Note that $\Pi_K=\varnothing$ for the \textbf{Vis-Only} baseline.
$\Pi_V\cup\Pi_K$ is converted into a factor graph, and the symbolic reasoning module executes its inference algorithm on the graph.
Marginal probabilities are queried afterwards, then the candidate concept with the highest probability is selected as answer.

\subsubsection{Explanation}

The explanation process is triggered when the agent parses a \textit{why}-\textsc{ques} like (\ref{eq:why_ques}) from the teacher, upon which it invokes a dedicated procedure for reflecting upon its reasoning process to select a \textit{sufficient reason} \cite{darwiche2020reasons}.
For example, if the agent knows the generic rule that fire trucks have ladders, then identifying a ladder with high confidence would be a sufficient reason for recognizing the whole object as a fire truck in our domain.
We implement a modified version of \citet{koopman2021persuasive}'s algorithm for finding sufficient reasons from our factor graphs.
We only consider the evidence variables in the factor graph (e.g., $ev\_dumper(p)$), which are observable, as potential explanans candidates.
If the algorithm returns `direct perceptions' as (part of) sufficient reasons, e.g., selecting $ev\_dumpTruck(o)$ as sufficient reason of the agent's answer $dumpTruck(o)$, they are deemed not informative and thus omitted from the explanation.
The intuition is that explanations like `I thought $o$ is a dump truck because it looked like a dump truck' do not reveal a knowledge gap that the teacher can exploit as a learning opportunity.

\subsubsection{Instance-level learning}

Instance-level learning updates the sets $\chi_\gamma^+$, $\chi_\gamma^-$ stored in the XB and thus contributes to updating $f_\text{clf}^\gamma$ and $f_\text{seg}^\gamma$.
As the sizes of $\chi_\gamma^{+/-}$ get larger, the accuracy of these functions should increase.  

$\chi_\gamma^+$ and $\chi_\gamma^-$ are updated whenever the agent makes an incorrect prediction.
Specifically, the teacher's correction ``This is not a dumper truck. This is a missile truck'' will add the referenced instance to $\chi_{dumpTruck}^-$ and $\chi_{missileTruck}^+$.
On any occasion where the agent made a correct prediction that an instance is a dump truck, but was not highly confident in that prediction, the instance is added to $\chi_{dumpTruck}^+$.
In this case, any recognized instances of relevant part concepts, as determined by the agent's current KB, are added to the XB as well; e.g., $\chi_{dumper}^+$ and $\chi_{quadCabin}^+$ would also be expanded in our example.
Note that the estimated binary masks for the part instances may be faulty and thus allow introduction of bad exemplars in the XB.
In fact, it is precisely these sources of error that we aim to mitigate via the teacher's corrective feedback in \textbf{Vis+Genr+Expl} dialogues.
That is, the corrective feedback ``This is not a quad cabin'' to the explanation ``Because I thought this is a quad cabin'' will add the misclassified non-instance to $\chi_{quadCabin}^-$.
This enables joint refinement of whole and part concepts.

\subsubsection{Rule-level learning}

For rule-level learning, the symbolic KB is updated whenever the teacher utters a generic statement after the learner fails to provide some verbal explanation for its incorrect prediction.
Each NL generic statement, like ``Dump trucks have dumpers'', is translated into a corresponding $\mathbb{G}$-quantified \textsc{prop} like (\ref{eq:genr_prop}) and added to the KB entry if not already included.
We assume the rules provided by the teacher are always correct.
KB rule addition is an idempotent operation in this work; we do not take any measure when already known rules are provided again.

\section{Experiments}
\label{sec:experiments}

\subsection{Setup}

We conduct a suite of experiments to assess the data-efficiency of the three different interaction strategies during FGVC training: \textbf{Vis-Only}, \textbf{Vis+Genr} and \textbf{Vis+Genr+Expl}.
We only compare against ablative baselines because no other existing neurosymbolic approaches are designed to cope with incremental learning combined with a lack of symbolic domain ontology after deployment.
We consider two FGVC domains with differing difficulties:
\begin{itemize}
    \item \textsf{single\_4way}: Four types of trucks can be distinguished solely by their load types.
    \item \textsf{double\_5way}: Five types of trucks can be distinguished by two dimensions of part properties (load and cabin), where some parts may be shared between truck type pairs and thus can fail to serve as distinguishing.
\end{itemize}
We also vary the part-recognition performance of the vision model with which the learner starts each experiment, by controlling initial XB entries, in order to study how the initial recognition capability affects the learning process.
We run experiments with three initial part recognition accuracies: \textsc{lq/mq/hq} (low-/medium-/high-quality).
Our evaluation metric is \textit{cumulative regret}; i.e., the accumulated number of mistakes made across a series of 120 interaction episodes, where subtypes and visual features are randomly sampled for each training instance.
We report cumulative regret curves averaged over 30 random seeds with 95\% confidence intervals.
Due to space limits, the Appendix offers more details on experiment setup and agent implementation.

\subsection{Results and discussion}

\begin{figure*}
\centering
\begin{subfigure}{0.23\textwidth}
    \includegraphics[width=\textwidth]{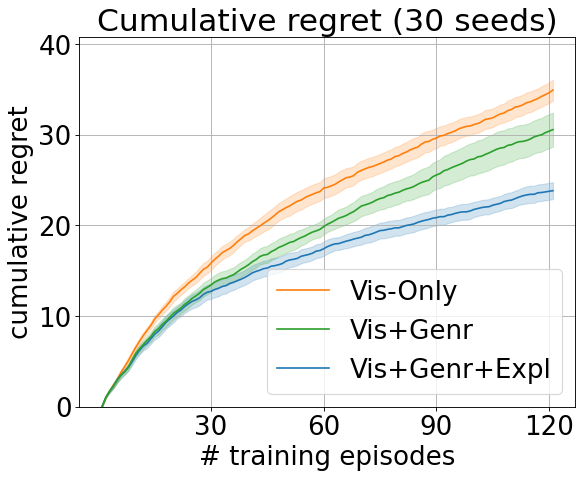}
    \caption{\textsf{single\_4way}, \textsc{lq}}
    \label{fig:cumul_regrets:4way}
\end{subfigure}
\begin{subfigure}{0.23\textwidth}
    \includegraphics[width=\textwidth]{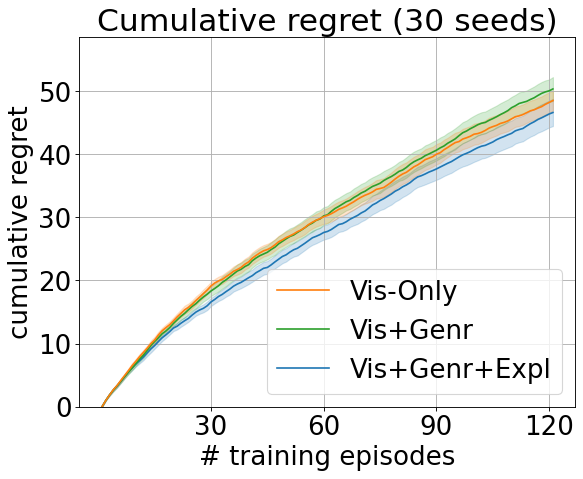}
    \caption{\textsf{double\_5way}, \textsc{lq}}
    \label{fig:cumul_regrets:5way_lq}
\end{subfigure}
\begin{subfigure}{0.23\textwidth}
    \includegraphics[width=\textwidth]{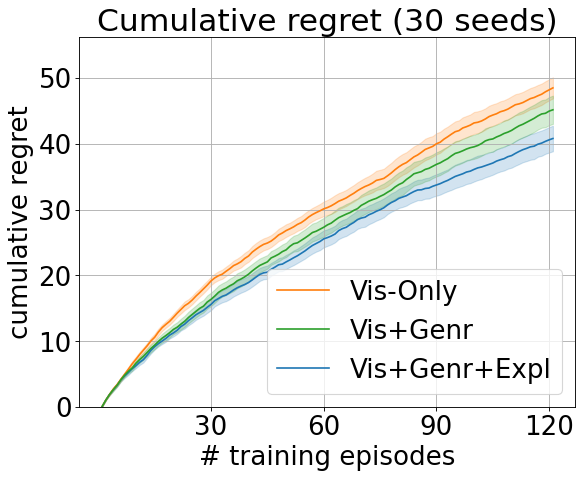}
    \caption{\textsf{double\_5way}, \textsc{mq}}
    \label{fig:cumul_regrets:5way_mq}
\end{subfigure}
\begin{subfigure}{0.23\textwidth}
    \includegraphics[width=\textwidth]{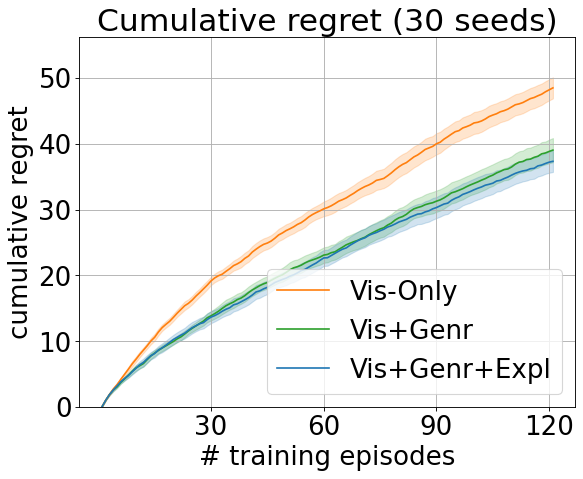}
    \caption{\textsf{double\_5way}, \textsc{hq}}
    \label{fig:cumul_regrets:5way_hq}
\end{subfigure}
\caption{Averaged cumulative regret curves (with 95\% confidence intervals): training example count vs. cumulative regret.}
\label{fig:cumul_regrets}
\end{figure*}

Plots in Fig.~\ref{fig:cumul_regrets} show how teacher-learner pairs adopting the three strategies acquire the target concepts at different rates.
Fig.~\ref{fig:cumul_regrets:4way} shows that the order for data efficiency of learning in \textsf{single\_4way} setting (starting from \textsc{lq}) is: \textbf{Vis+Genr+Expl}, \textbf{Vis+Genr}, \textbf{Vis-Only}, from the most to the least efficient.
Two-sample $t$-tests on their final cumulative regret values show these differences are all statistically significant ($p<10^{-6}$), also attested by the wide separations between the 95\% confidence intervals in Fig.~\ref{fig:cumul_regrets:4way}.
Generic rules allow the learner to generalize better by attending to more important visual features (e.g., part types) while mitigating distractions by irrelevant features (e.g., colors).
Teacher feedback to learner explanations provides even further synergies through improving part recognition models.

The results for the \textsf{double\_5way} experiments are nuanced.
In particular, notice in Fig.~\ref{fig:cumul_regrets:5way_lq} how \textbf{Vis+Genr} performs worse than \textbf{Vis-Only}.
This may be due to the fact that the cabin parts (quad vs. hemtt) are more difficult to tell apart: they look more similar to each other and have varying colors as distracting features.
This suggests that depending on task difficulty, generic part-whole rules may actually cause an adverse effect on the learning progress if recognition capabilities for auxiliary feature concepts are too inaccurate and not properly refined during learning.
\textbf{Vis+Genr+Expl} manages to take back the (marginal) comparative advantage in data efficiency by joint refinement of concept boundaries for both object parts and wholes.

Fig.~\ref{fig:cumul_regrets:5way_mq} and Fig.~\ref{fig:cumul_regrets:5way_hq} show that when we start with better part recognition models---namely, the \textsc{mq} and \textsc{hq} models---then generic rules improve data efficiency again.
Notice how \textbf{Vis+Genr+Expl} consistently outperforms \textbf{Vis+Genr} with all \textsc{lq}, \textsc{mq} and \textsc{hq} models, though the confidence intervals start to overlap in Fig.~\ref{fig:cumul_regrets:5way_hq}.
Two-sample $t$-tests on final regret values between \textbf{Vis+Genr} vs. \textbf{Vis+Genr+Expl} with \textsc{lq/mq/hq} as starting model yield $p$-values of 0.015/0.005/0.187 respectively.
This makes sense, considering the fact that the \textsc{hq} part model already has strong part recognition capability and leaves relatively little room for improvement.
Overall, one needs generics combined with explanations to \textit{guarantee} that improving ontological knowledge enhances data efficiency in visual tasks, regardless of the quality of the initial vision model.

\section{Related Work}
\label{sec:background}

A line of research investigates a framework often referred to as explanatory interactive learning (XIL), which focuses on the corrective role of explanations in teacher-learner interactions \cite{teso2023leveraging}.
\textsc{caipi} \cite{teso2019explanatory} is a model-agnostic method that converts local explanations to counterexamples, but it cannot analytically state how agent explanations should be corrected at the feature level.
\citet{schramowski2020making} extend \textsc{caipi} by allowing users to interact with explanations given as input gradients at the expense of model-agnosticity.
ProtoPDebug \cite{bontempelli2023concept} makes explicit references to visual parts for human-in-the-loop debugging of neural FGVC models, but not in the form of logical rules.
NeSy XIL \cite{stammer2021right} bears a strong resemblance to our approach in that it implements a neurosymbolic architecture which admits rule-based correction from local explanations.
However, NeSy XIL heavily relies on the traditional DNN training scheme and thus is more suitable for static task domains where human inspections are needed only occasionally.
\citet{michael2019machine} and \citet{tsamoura2021neural} are another relevant line of work with a heavy emphasis on neurosymbolic architectures and argumentation-based model coaching, but they are also chiefly concerned with guiding parameter updates of neural models in static domains.
HELPER \cite{sarch2023open} is another closely relevant approach to building customizable agents with memory-augmented large neural models, but it is designed to learn novel user-specific action routines, while our approach focuses on enabling a pretrained neural model to acquire concepts it was completely unaware of.

Interactive task learning (ITL; \citealt{laird2017interactive}) is a ML paradigm motivated by scenarios where an AI system needs to cope with unforeseen changes after deployment.
Learning in ITL takes place primarily through natural interactions with a human domain expert.
Therefore, it is natural for ITL to utilize insights from linguistic phenomena, such as discourse coherence \cite{appelgren2020interactive} and complex referential expressions \cite{rubavicius2022interactive}.
We contribute to this body of work by extending the neurosymbolic ITL architecture proposed in \citet{park2023interactive}, which exploits the truth-conditional semantics of generic statements; our extension adds corrective feedback to the agent's explanations as further learning opportunities.

\section{Conclusion and Future Directions}
\label{sec:conclusion}

In this paper, we have proposed an explanatory interactive learning framework for neurosymbolic architectures designed to solve knowledge-intensive tasks.
It uses natural language dialogues through which the teacher provides feedback to the learner's explanations of its (incorrect) predictions.
This feedback provides piecemeal information of domain ontologies, enabling incremental refinement of imperfect visual concept boundaries as and when needed.
Symbolic reasoning enhances generalizability by explicitly highlighting important feature concepts, while correction of inaccurate part-based explanations ensures the referenced visual evidence is of good quality.
Our proof-of-concept experiments show that timely exploitation of explanations can significantly boost learning efficiency in tasks where knowledge of part-whole relations is integral, although such improvements may be conditional upon the quality of the part recognition model.
Potential future research directions include: in-depth exploration of exception-admitting generics \cite{pelletier1997generics}, extension to continuous regression tasks \cite{letzgus2022toward} and long-horizon planning tasks \cite{fox2017explainable}.

\section*{Acknowledgements}

This work was supported by Informatics Global PhD Scholarships, funded by the School of Informatics at The University of Edinburgh.
Ramamoorthy is supported by a UKRI Turing AI World Leading Researcher Fellowship on AI for Person-Centred and Teachable Autonomy (grant EP/Z534833/1).
We thank the anonymous reviewers for their feedback on an earlier draft of this paper, and Rimvydas Rubavicius and Gautier Dagan for continued feedback over the course of this research.

\bibliography{aaai25}

\appendix

\section{Technical Appendix}

\subsection{Vision processing module: \\Detailed formal specifications}

Formally, suppose the vision module is provided with a raw RGB image input with width $W$ and height $H$ as $\mathcal{I}\in[0,1]^{3\times H\times W}$.
The agent's current vocabulary of unary and binary visual concepts are represented with sets $C_u$ and $C_b$ respectively.
We do not make any exclusivity assumption among concepts, so $C_u$ and $C_b$ can contain concepts that are supertypes or subtypes of others.
We represent a scene graph containing $N$ objects as a pair of a vertex set and an edge set $SG=(V,E)$ such that:
\begin{itemize}
    \item $V=\{(\mathbf{c}_i,m_i)\mid i\in [1..N]\}$, where $\mathbf{c}_i\in[0,1]^{|C_u|}$ represents the probabilistic beliefs of whether the object indexed by $i$ classifies as an instance of unary concepts in $C_u$, and $m_i\in\{0,1\}^{H\times W}$ a binary segmentation mask referring to the object.
    \item $E=\{\mathbf{r}_{i,j}\mid i,j\in [1..N], i\neq j\}$, where $\mathbf{r}_{i,j}\in[0,1]^{|C_b|}$ represents the probabilistic beliefs of whether the ordered pair of objects indexed by $(i,j)$ classifies as an instance of binary concepts in $C_b$.
\end{itemize}
$SG$ serves as an internal, abstract representation of $\mathcal{I}$ that will be later processed by symbolic reasoning methods.

The functional requirement we impose on the vision processing module is twofold.
First, given a raw RGB image $\mathcal{I}\in[0,1]^{3\times H\times W}$ and an ordered set of $n$ binary masks $m_1,\cdots,m_n\in\{0,1\}^{H\times W}$ referring to scene objects, the vision module should be able to perform few-shot binary classification on known visual concepts, returning the estimated probabilities of concept membership as its output.
That is, if we denote sets of positive and negative examples of a visual concept $\gamma$ as $\chi_\gamma^+$ and $\chi_\gamma^-$ respectively, the vision module should provide a function $f_\text{clf}^\gamma$ for each concept $\gamma$ such that $f_\text{clf}^\gamma(\mathcal{I},(m_1,\cdots,m_n),(\chi_\gamma^+,\chi_\gamma^-))$ returns the estimated probability that the $n$-tuple of objects referenced by $(m_1,\cdots,m_n)$ is an instance of $\gamma$.
We steer clear of the closed-world assumption by estimating a binary distribution for each individual concept instead of a multinomial distribution over a fixed set of concepts, accommodating incremental learning of an open vocabulary of concepts.
In our setting, we need $f_\text{clf}^\gamma$ defined for $n=1$ (for $C_u$) and $n=2$ (for $C_b$); the former estimates $\mathbf{c}_i$ in $V$ for each object $i$ in the scene, the latter estimates $\mathbf{r}_{i,j}$ in $E$ for each object pair $(i,j)$.

Second, given a raw RGB image $\mathcal{I}\in[0,1]^{3\times H\times W}$ and a visual concept $\gamma$ specified by a set of positive concept examples $\chi_\gamma^+$, the vision module should be able to localize and segment instances of $\gamma$ in the visual scene as binary masks $m_1,\cdots,m_p\in\{0,1\}^{H\times W}$ for some $p$.
This `search' functionality is needed for recognizing object parts as evidence that would significantly affect later symbolic reasoning.
As with the classification function $f_\text{clf}^\gamma$, the segmentation function $f_\text{seg}^\gamma$ is required to be few-shot as well.
The set of $p$ proposal masks returned by $f_\text{seg}^\gamma(\mathcal{I},\chi_\gamma^+)$ are fed into $f_\text{clf}^\gamma$ for re-evaluation (with $\chi_\gamma^-$ also provided as needed), from which the top-scoring proposal(s) can be selected as best match and added to the scene graph.
In our setting, we are only interested in searching for instances of unary concepts in $C_u$, object part concepts in particular.

\subsection{Further implementational details \\on experiment setup}

\subsubsection{Evaluation scheme}

As mentioned in the main text, our suite of experiments is designed to assess the data-efficiency of the three different interaction strategies during FGVC training: \textbf{Vis-Only}, \textbf{Vis+Genr} and \textbf{Vis+Genr+Expl}.
We consider two FGVC domains with differing difficulties:
\begin{itemize}
    \item \textsf{single\_4way}: Four types of trucks can be distinguished solely by their load types.
    \item \textsf{double\_5way}: Five types of trucks can be distinguished by two dimensions of part properties (load and cabin), where some properties may be shared between truck type pairs and thus can fail to serve as distinguishing.
\end{itemize}
Each strategy is evaluated by exposing the learner to sequences of interaction episodes as delineated in Fig.~2 in the main text.
We use 30 different sequences of interaction episodes per domain, where each sequence consists of 120 training examples sampled by randomizing both relevant and distracting features.
For relevant features, we vary the subtypes of the truck load and cabin subparts.
For distracting features, we vary colors of parts, number and size of wheels, and position and orientation of trucks (see Fig.~1a in the main text).
A fixed set of example sequences from 30 random seeds are shared across the strategies for controlled sampling.

The learner starts each episode sequence with zero knowledge of the domain ontology and a deficient model for recognition of parts.
The quality of recognition model for each part concept $\gamma$ can be controlled by the sizes of $\chi_\gamma^+$ and $\chi_\gamma^-$ in the visual XB.
We indirectly control sizes of the part exemplar sets by running a prior training sequence in \textbf{Vis-Only} mode with varying numbers of episodes.
The more part examples the agent has witnessed before the interactive FGVC task, the larger $\chi_\gamma^+$ and $\chi_\gamma^-$ it will have, and so the higher the quality of the part recognition model is.
The prior part models we used in our experiments yielded averaged classification accuracy of 74.83\%/88.86\%/98.17\% after 20/100/200 part training episodes.
We will refer to each of them as \textsc{lq/mq/hq} (low-/medium-/high-quality) models.

Our evaluation metric is \textit{cumulative regret}; i.e., the accumulated number of mistakes made across each 120-episode sequence.
We want to minimize cumulative regret (at each step in the sequence) in order to claim better training data-efficiency.
This choice of evaluation metric reflects the nature of our interactive learning framework, in which training and inference are closely intertwined, and each piece of user feedback should take effect right after it is provided.
We report cumulative regret curves averaged over 30 seeds with 95\% confidence intervals.

\subsubsection{Implementation}

We randomly generate visual scene images from a simulated environment implemented in Unity.
In the \textsf{single\_4way} domain, $C_\text{fg\_type}\stackrel{\text{def}}{=}C_\text{4way}=\{\text{\textit{baseTruck, dumpTruck, missileTruck, fireTruck}}\}$.
For \textsf{double\_5way}, $C_\text{5way}=C_\text{4way}\cup\{\text{\textit{containerTruck}}\}$.
See Fig.~1a in the main text for the distinguishing part properties of each truck type.
The teacher only refers to generic rules about load types in \textsf{single\_4way}, while rules about both load and cabin types are taught in \textsf{double\_5way}.
We implement simulated teachers in our experiments that can respond to the learner's NL utterances, just so that the dialogue flows described in Fig.~2 in the main text are correctly followed.

We implement our neurosymbolic architecture by employing existing technology stacks that are available off-the-shelf or readily extensible.
For the vision module, we use DINOv2-Small \cite{oquab2023dinov2} for patch- and instance-level feature extraction, and SAM \cite{kirillov2023segment} for instance segmentation.
$f_\text{clf}^\gamma$ functions are implemented as a distance-weighted kNN classifier on \texttt{[CLS]}-pooled vector outputs from DINOv2, which are obtained by feeding visual-prompted image segments \cite{yang2024fine}.
$f_\text{seg}^\gamma$ functions are implemented by coordinating DINOv2 and SAM as done in Matcher \cite{liu2024matcher}, a training-free framework for few-shot instance segmentation.
The language processing module employs a large-coverage parser of the English Resource Grammar \cite{copestake2000open}, whose outputs are heuristically translated into our \textsc{prop} and \textsc{ques} formalisms.
For symbolic reasoning, we extend the loopy belief propagation algorithm \cite{murphy1999loopy} for approximate inference, modified to comply with the semantics of logic programs.

\end{document}